\documentclass[journal]{IEEEtran}
\usepackage{amsmath,amsfonts}
\usepackage{algorithmic}
\usepackage{algorithm}
\usepackage{array}
\usepackage[caption=false,font=footnotesize,labelfont=rm,textfont=rm]{subfig}
\usepackage{textcomp}
\usepackage{stfloats}
\usepackage{url}
\usepackage{verbatim}
\usepackage{graphicx}
\usepackage{cite}
\usepackage{booktabs}
 \usepackage{color} 
\usepackage{makecell}
\usepackage{stfloats}
\usepackage[colorlinks=true,urlcolor=red,citecolor=blue,linkcolor=blue]{hyperref}
\usepackage[section]{placeins}
\usepackage{amssymb}
\usepackage{indentfirst}
\usepackage{subfig}
\DeclareUnicodeCharacter{E011}{\ensuremath{-}}
\usepackage[marginal]{footmisc}
\usepackage{multirow}
\usepackage{color}
\usepackage{fancyhdr}

\makeatletter 
\let\myorg@bibitem\bibitem
\def\bibitem#1#2\par{%
  \@ifundefined{bibitem@#1}{%
    \myorg@bibitem{#1}#2\par
  }{%
    \begingroup
      \color{\csname bibitem@#1\endcsname}%
      \myorg@bibitem{#1}#2\par
    \endgroup
  }%
}
\makeatother %

\begin{document}

\title{FedForgery: Generalized Face Forgery Detection with Residual Federated Learning}

\author{Decheng~Liu,~Zhan~Dang,~Chunlei~Peng,~\IEEEmembership{Member, IEEE}~,~Yu~Zheng,~Shuang~Li,~Nannan~Wang,~\IEEEmembership{Member, IEEE}~and Xinbo Gao, \IEEEmembership{Senior Member, IEEE}
\noindent

\thanks{\indent D. Liu, Z. Dang, and C. Peng are with the State Key Laboratory of Integrated Services Networks, School of Cyber Engineering, Xidian University, Xi’an 710071, Shaanxi, P. R. China and with Shanghai Key Laboratory of Computer Software Evaluating and Testing, Shanghai 201112, P. R. China (e-mail: dchliu@xidian.edu.cn; zd.xidian@gmail.com; clpeng@xidian.edu.cn).\\
\indent Y. Zheng is with the School of Cyber Engineering, Xidian University, Xi’an 710071, Shaanxi, P. R. China. (e-mail: yzheng@xidian.edu.cn).\\
\indent S. Li is with Shanghai Key Laboratory of Computer Software Evaluating and Testing, Shanghai 201112, P. R. China. (e-mail: ls@sscenter.sh.cn).\\
\indent N. Wang is with the State Key Laboratory of Integrated Services Networks, School of Telecommunications Engineering, Xidian University, Xi’an 710071, Shaanxi, P. R. China (e-mail: nnwang@xidian.edu.cn).\\
\indent X. Gao is with the Chongqing Key Laboratory of Image Cognition, Chongqing University of Posts and Telecommunications, Chongqing 400065, P. R. China.(e-mail: gaoxb@cqupt.edu.cn).}
}

\markboth{~Vol.~14, No.~8, October~2022}%
{IEEE Transactions on Information Forensics and Security}


\maketitle
\thispagestyle{fancy}
\cfoot{\small{\copyright~2023 IEEE. Personal use of the material is permitted. 
Permission from IEEE must be obtained for all other uses, in any current or future media, including reprinting/republishing this material for advertising or promotional purposes, creating new collective works, for resale or redistribution to servers or lists, or reuse of any copyrighted component of this work in other works.}}
\rfoot{}

\begin{abstract}
With the continuous development of deep learning in the field of image generation models, a large number of vivid forged faces have been generated and spread on the Internet.
These high-authenticity artifacts could grow into a threat to society security.
Existing face forgery detection methods directly utilize the obtained public shared or centralized data for training but ignore the personal privacy and security issues when personal data couldn’t be centralizedly shared in real-world scenarios.
Additionally, different distributions caused by diverse artifact types would further bring adverse influences on the forgery detection task.
To solve the mentioned problems, the paper proposes a novel generalized residual \textbf{Fed}erated learning for face \textbf{Forgery} detection (\textbf{FedForgery}).
The designed variational autoencoder aims to learn robust discriminative residual feature maps to detect forgery faces (with diverse or even unknown artifact types).
Furthermore, the general federated learning strategy is introduced to construct distributed detection model trained collaboratively with multiple local decentralized devices, which could further boost the representation generalization.
Experiments conducted on publicly available face forgery detection datasets prove the superior performance of the proposed FedForgery.
The designed novel generalized face forgery detection protocols and source code would be publicly available at  \emph{https://github.com/GANG370/FedForgery.}
\end{abstract}

\begin{IEEEkeywords}
Facial forgery detection, residual feature learning, federated learning, privacy preserving.
\end{IEEEkeywords}

\section{Introduction}
\IEEEPARstart{W}{ith} the large development of deep generative models, face forgery technology has been spread widely and rapidly. Consequently, the artifact forgery clue has become more and more difficult to identify, even undistinguishable to human eyes. Meanwhile, the forgery technology may be abused by criminals, by changing the faces of public stars and politicians’ videos, thereby spreading some negative public information; attacking the face identification system in the transportation hub, impersonating target identities for pursuit-evasion, etc. 
There is no doubt that these dangerous behaviors are very likely to bring about unpredictable effects on social security.
Although a lot of researchers have been focused on face forgery detection tasks, these artifact videos contain large amounts of redundant data, complex backgrounds, and diverse artifact types, which would inevitably bring negative effects on forgery detection tasks. Thus, face forgery detection is still a challenging and important problem in real-world scenarios.

Face forgery detection can be easily defined as a binary classification problem.
Existing forgery detection methods can be roughly divided into two categories: forgery image detection methods and forgery video detection methods. 
The forgery image detection methods mainly focus on the difference between the real and the fake images in the low-level information, such as utilizing the image frequency domain information \cite{chen2021local} and \cite{qian2020thinking}, the image mixed boundary information \cite{li2020face}, and the extra head pose features \cite{yang2019exposing}, etc. Different from forgery image detection, forgery video detection methods mainly perform authenticity detection based on the visually unnatural image cross-frame transition and temporal inconsistency of fake videos. 
For example, \cite{li2020sharp} utilized a multi-instance learning framework to capture the temporal inconsistency of faces for deep fakes detection. Sabir et al. \cite{sabir2019recurrent} proposed the convolution recursive model to extract the temporal information of the video stream to detect the authenticity of the video. 
It is noted that existing detection methods usually required centralized training data for training model parameters, which makes it easy to leak personal privacy.
How to effectively train models with distributed training data on different devices and protect personal privacy draws more and more attention in recent years.
In addition, different forgery distributions caused by diverse artifact types would further bring adverse influences.
\emph{To mimic the real-world scenarios, we newly design two novel face forgery detection tasks: Hybrid-dataset forgery detection and Generalized-dataset forgery detection task as shown in Figure 1. 
}
\begin{figure*}[ht]
    \centering
    \includegraphics[height=7.4cm, width=17.5cm]{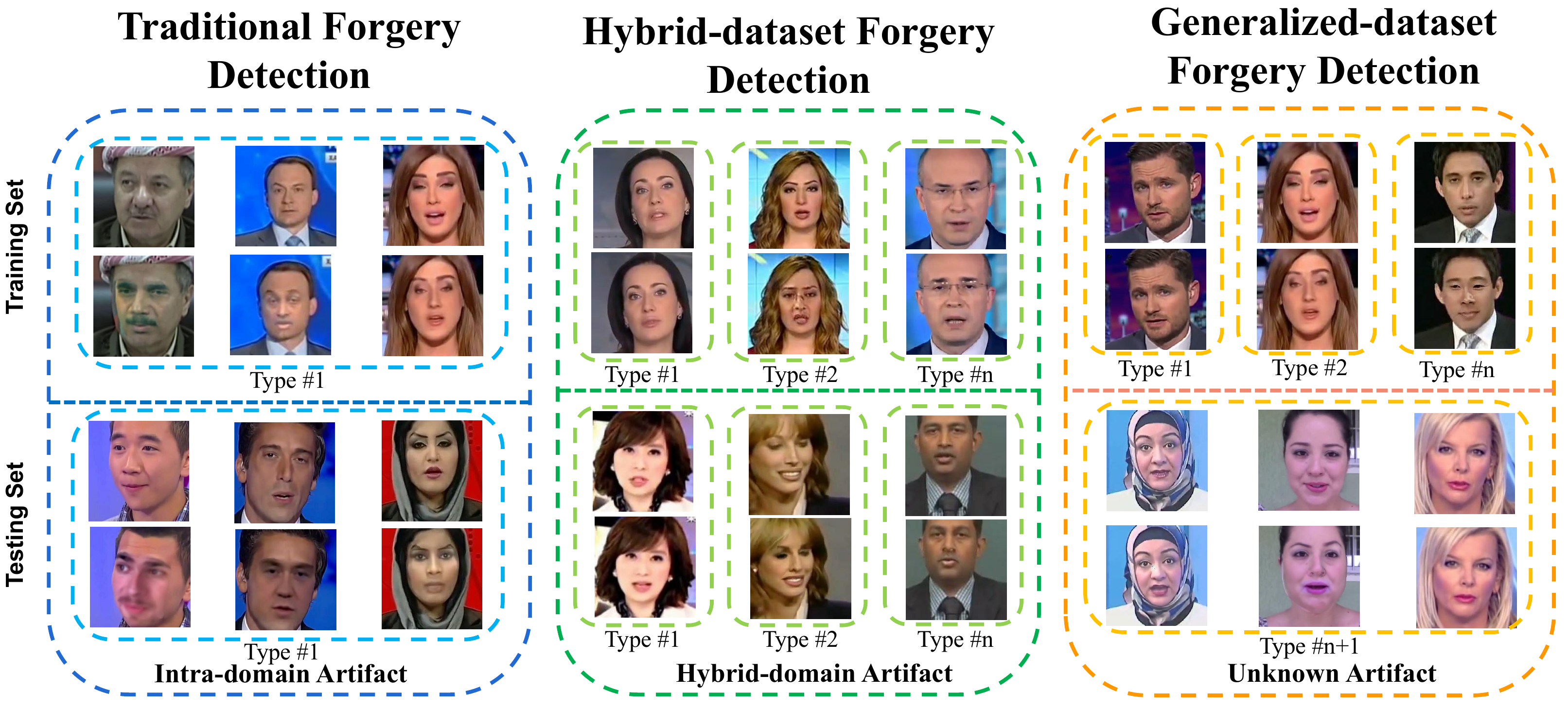}
    \caption{\textbf{Problem setup.} The left subfigure shows the traditional forgery detection task. The same artifact type forgery dataset is selected in both the training and testing stages.
    The middle subfigure shows the designed Hybrid-dataset forgery detection task, where multiple artifact types are mixed as the training data and the evaluation model aims to distinguish the authenticity of input faces.
    The right subfigure shows the designed Generalized-dataset forgery detection task. Different from the middle task, the goal of this task is to distinguish the real from the fake even for unknown artifact types.
    }
\end{figure*}

Recently, federated learning has been developed rapidly and widely deployed in the real world \cite{mcmahan2017communication}. 
In the federated learning architecture, each local client could only use local private data for training the local model firstly, and then upload the model parameters to the global server for training the global model. 
After aggregating the model parameters several times, the server model parameters will be sent to each data center for updating. The aggregation update process will be iterative until the training loss converges. 
The federated learning strategy inspires us to design a suitable training method to protect local data privacy, and further learn robust forgery discriminative features with strong generalization ability.

This paper proposes a novel generalized residual Federated learning for face Forgery detection (FedForgery), which not only could train a global model collaboratively for security restrictions but also would improve the discriminative forgery representation generalization ability.
Firstly, considering diverse artifact distribution in the generalized forgery detection tasks, we design the variational autoencoder model to learn robust residual feature maps for forgery clues.
Secondly, the proposed residual federated learning algorithm is introduced to update distributed local model parameters and  global server model parameters collaboratively, which could further boost the forgery representation generalization ability. 
Finally, the global model in the server could directly detect different artifact types and even unknown type forgery.
\emph{The main reason for superior performance is that we effectively propose the residual feature map for generalized forgery detection, and simultaneously introduce a specific federated learning strategy to boost performance and protect data privacy.}

To the best of our knowledge, it is the first exploration to introduce federated learning and explore generalization ability in the face forgery detection field. The main contributions of our paper can be summarized as follows:
\begin{enumerate}
\item We firstly explore a residual federated learning framework for face forgery detection, which could effectively construct a distributed model with multiple local decentralized devices, and also boost the representation generalization ability.
  
\item The designed variational autoencoder is proposed to analyze the difference of reconstruction residuals between the real and fake images in diverse artifact types, which can help representation identify hybrid and even unknown artifact types.

\item Experimental results on several public face forgery detection datasets illustrate the superior performance of the proposed FedForgery compared with state-of-the-art methods. The designed protocols and code would be publicly available at  \emph{https://github.com/GANG370/FedForgery.}
\end{enumerate}

We organized the rest of this paper as follows. Section I gives a brief introduction to the proposed method, and Section II shows some representative face forgery detection algorithms. In Section III, we present the novel generalized face forgery detection with residual federated learning for face forgery detection. Section IV shows the experimental results and analysis. Section V shows the ablation results and visualized analysis. The conclusion is drawn in Section VI.
\section{Related Work}
\subsection{Face Forgery Detection}
\label{A.Fake face detection}

In recent years, with the increasing popularity of face forgery technology, research on face forgery detection has gradually developed. 
Face forgery detection is essentially a binary classification problem, the purpose of it is to train a classifier with a high degree of precision and generalizability. 
As far as the currently proposed methods are concerned, they can be divided into image-based face forgery detection methods and video-based face forgery detection methods. 

There are currently methods to detect forgery images with the help of image-level labels. 
For example, Chen \emph{et al.} \cite{chen2021local} proposed a general method that utilizes the RFAM module to fuse image RGB domain information and frequency domain information to obtain a more comprehensive local feature representation, so as to conduct face image authenticity detection. 
The x-ray surveillance system \cite{li2020face} is designed  to detect deepfakes with the artifact clues generated by mixing two images.
Nirkin \emph{et al.} \cite{nirkin2021deepfake} described a method using the proposed face recognition network and background recognition network. 
Zhou \emph{et al.} \cite{zhou2017two} designed a two-stream network to detect low-level inconsistencies between tampered faces and image patches in order to distinguish fake face images. 
Yang \emph{et al.} \cite{yang2019exposing} observed that these deepfake images are generated by stitching synthesized face regions into original images, thus the novel method is proposed to detect fake videos by inputting 3D head pose features into support vector machines. 
Li \emph{et al.} \cite{li2018exposing} utilized the designed convolutional neural network to capture the artifacts in fake images to identify the authenticity of the image. 
Yu \emph{et al.} \cite{9694644} designed the novel U-net structure and several designed losses to train a universal forgery feature extractor, which aims to explore robust forgery traces and learn a better generalization ability when detecting unknown forgery types.
Scherhag \emph{et al.} \cite{9093905} proposed a differential MAD algorithm based on deep face representation to extract rich and compact deep facial representations from pairs of reference and probe images in combination with training a machine learning-based classifier to detect image changes caused by the morphing algorithm.
Yang \emph{et al.} \cite{9505637} studied a multi-scale texture difference model MTD-Net, which uses a special convolution operation of central difference convolution for robust face forgery detection. This is the first attempt to introduce a special convolution operation for feature extraction and information fusion in the field of face forgery detection.
In order to adapt to insufficient annotated data, Zhao  \emph{et al.} \cite{2017Marginalized} proposed the mCNN to learn invariant representation to improve the representation generalization in face analysis task.
Miao \emph{et al.} \cite{9854878} pointed out that a new hierarchical frequency-aided interaction network for face forgery detection, which utilizes a frequency-based feature refinement module to extract mid-high frequency traces on rgb features, making full use of more general image frequency domain features.
Wang \emph{et al.} \cite{9808159} designed a new localization Invariant Siamese network, which uses the localization invariant loss to improve the localization consistency between two segmentation maps, so as to strengthen the localization invariance to different image degradations, and to perform more efficient deepfake detection.    
Wang \emph{et al.} \cite{9824499} proposed an attention mechanism to capture local artifact features from facial attention regions, and allows multiple attention maps to focus on different regions of the face, thereby improving forgery detection performance.
The method based on frequency domain analysis is widely used in image analysis, image reconstruction and other image processing.
Durall \emph{et al.} \cite{durall2019unmasking} used the DFT transformations to extract features and then send them to a classifier for training. 
Dong \emph{et al.} \cite{dongprotecting} proposed the novel identity consistency transformer to find identity inconsistency in face regions, which also exhibits its good generalization ability in real-world applications.
Zhu \emph{et al.}\cite{zhuface} firstly disentangled a face into four graphic components with the designed 3D decomposition method, and then utilized the composition search strategy to mine forgery clues from relevant components.
Zhao \emph{et al.}\cite{zhaomulti} proposed a novel multi-attention deepfake detection network, which incorporates the designed region-independence loss and an attention-guided data augmentation strategy to improve performance.
Different from  Cao \emph{et al.}  \cite{cao2022end} distinguished  between real faces and fake faces by learning a common compact representation of real faces, the proposed method exploits the similarity between the reconstructed residuals of different forged types of data to improve generalization. 
However, these image-based methods need to collect all data from the data center during model training, but could not be applied in distributed storage non-public video data scenarios.

Video-based methods are mainly based on the detection of visually unnatural image cross-frame transitions and temporal inconsistencies in fake videos. 
Li \emph{et al.} \cite{li2020sharp} proposed a multi-instance learning framework to treat faces and input videos as instances and bags respectively. 
They further utilized a sharp mapping method to map instance embeddings to bag prediction. 
Sabir \emph{et al.} \cite{sabir2019recurrent} studied the novel convolutional recurrent models to extract features in terms of temporal information of image streams to detect fake data.
CIFTCIUA \emph{et al.}\cite{ciftciua2020detection} found that the authenticity of the videos can be detected by biological signals. From this point, a video classifier based on the synthesis of physiological signal changes was created. However, getting a heart rate signal from the video is not an easy job.
Saikia \emph{et al.}\cite{saikia2022hybrid} described a CNN model based on optical flow volume, pre-trained CNN model and LSTM layer to model the inconsistent motion of each pixel in a video frame, so as to distinguish the true and false video.
Masi \emph{et al.}\cite{masi2020two} designed a deep fake detection method based on a two-branch network structure, one branch is used to propagate the original information, and the other branch is used to suppress the face content. In the later stage of the network, a bidirectional long short-term memory architecture is used to learn the temporal features between video sequences.
Ganiyusufoglu \emph{et al.}\cite{ganiyusufoglu2020spatio} proposed to use 3D convolutional neural networks to model spatio-temporal features to capture the local spatio-temporal relationships and inconsistencies of deepfake videos, so as to extend the generalization function to detect new types of deepfake videos.
Trinh \emph{et al.}\cite{trinh2021interpretable} designed to consider unnatural motion and temporal artifacts as a form of visual interpretation, learn temporally inconsistent prototype representations in the latent space by combining spatial and temporal information, and then make predictions based on the similarity between the dynamic prototype of the test video and the learned dynamic prototype.
Guera \emph{et al.}\cite{guera2018deepfake} studied a time-aware convolutional neural network that created sequence descriptors and a fully connected layer to automatically detect deep fake videos.
Gu \emph{et al.}\cite{gu2021spatiotemporal} designed a spatial-temporal inconsistency learning module, which takes advantage of the temporal differences between adjacent frames along horizontal and vertical directions. The inconsistencies between single and consecutive frames in deepfake videos are jointly learned to obtain a more comprehensive representation.
In \cite{li2018ictu}, the image sequence of the eye region was exacted and the LSTM network was utilized to predict the blink probability to determine the authenticity. 
The limitation of this method is that the person in the video must be in a state of open eyes.
The new spatiotemporal convolution method \cite{de2020deepfake} is proposed to detect artifact traces across video frames for video authenticity identification. Video-based detection methods are always computationally expensive compared with image-based methods, because of fusing temporal inconsistencies.
Overall, most image-based forgery detection and video-based forgery detection methods both ignore the issue of data privacy.
They always utilized shared or centralized images when training the model, which increases the risk of data leakage. 
In addition, limited research has explored the distributed storage of training data in the forgery detection field.
\subsection{Federated Learning }
\label{B.Federated learning }
Federated learning could help protect data privacy and has drawn more and more attention recently.
There always exist two components: data center and global server in a typical federated learning framework. 
Each local data center only uploads parameters to update the global server model with a specific aggregated strategy. 
McMahan \emph{et al.} \cite{mcmahan2017communication} firstly introduced the idea of the federated learning pipeline.
The distinctive feature of the algorithm is the weighted average of model parameters on the global server. The FedProx algorithm proposed by Li \emph{et al.}\cite{li2020federated} solves the problem of data heterogeneity distributed across systems and networks in federated learning, and significantly improves the convergence behavior of federated learning with heterogeneous networks in the real world.

The privacy-preserving performance of federated learning has received more and more attention. 
To protect sensitive personal information in person re-identification tasks, a distributed federated learning method with unlabeled data is adopted in \cite{zhuang2021joint} for cloud and edge federated optimization. 
Yao \emph{et al.} \cite{yao2022federated} proposed the DualAdapt model for multi-objective domain adaptation tasks in image classification and semantic segmentation, which utilized federated learning distributed framework to deal with the domain gap between unlabeled data on the client and labeled centralized datasets on the server. 
Shao \emph{et al.}\cite{shao2022federated} studied that common face representation can be detected by introducing a federated domain decoupling strategy. This is the first to study the federated learning technique for the task of face presentation attack detection. 
Zhou \emph{et al.}\cite{9772495} designed to apply federated learning to edge computing. By designing a flexible participation training mechanism, it achieved low communication overhead and protected the privacy of edge clients, which benefited edge computing.
Liu \emph{et al.}\cite{9524709} proposed a privacy-enhanced based federated learning framework using homomorphic encryption as the underlying technique, which is the early exploration to effectively detect poisoning behavior in the federated learning process.
Niu \emph{et al.}\cite{niu2022federated} focused on the problem of unconstrained face recognition with private decentralized data, and they proposed a softmax-based regularization method to revise the class embedding gradients to enhance the discriminative power of class embedding across clients.
Zhuang \emph{et al.}\cite{zhuang2022federated} studied a cluster-based domain-adaptive federated learning method to improve the recognition performance of the target domain, addressing the problem of face recognition under privacy constraints. 
However, limited works applied federated learning in face forgery detection tasks.
The number of training data has grown rapidly, and the issue of data privacy protection becomes more and more important.
Besides, the distributed data collaborative training strategy also could help improve the model to avoid overfitting.
Considering these issues, the paper proposes a novel generalized residual federated learning for face forgery detection to solve these problems.

\section{Proposed Approach}
In this section, we will give details of the proposed FedForgery. 
 We analyze the residual feature maps between input images and reconstruction images under different artifact types.
 The designed variational autoencoder module could help identify hybrid-dataset artifacts and even unknown artifact types. 
 The overall framework of FedForgery is shown in Figure 2.
 The face forgery images are distributed storage. 
 $K$ data centers possess $K$ private different datasets: $D$$_{1}$, $D$$_{2}$, ... $D$$_{K}$. 
 Each data center only performs local training processing with local private data. 
And then, the local face forgery detection model parameterized by $w$ is obtained.

\subsection{Discriminative Residual Feature Learning}
\label{A. Image reconstruction represents learning }
Related researches \cite{cao2022end} \cite{jiang2020deeperforensics}  have proved that different data distribution could bring adverse effects on forgery detection performance, especially in defined Hybrid-dataset and Generalized-dataset forgery detection scenarios in Section I. Due to the diverse forgery artifact types, these existing domain gaps even may be larger than the difference between real and fake images. It makes it still a challenging problem to distinguish the authenticity in unknown artifact scenarios. Thus, we assume that exploring the specific characteristics of reconstruction residual is easier than inputting raw images in the multiple domain forgery scenarios. It also inspires us to design suitable robust residual features learning to avoid overfitting and effectively capture domain-invariant discrepancy information between real and fake images.
As shown in Figure 2, each data center aims to train a local face forgery detection model first.
We design a variational autoencoder to analyze the difference between real images and artifact reconstruction residuals under different artifact types. 
The local client model consists of two networks: RecNet represents the face reconstruction network, and ClsNet represents the forgery classification network. 
In the RecNet module, we utilized the variational autoencoder $G(\cdot )$ with similar architecture \cite{van2017neural} to reconstruct input face $x$ to $G(X)$.
Then we calculate the residual $x-G(X)$ between the original image and the reconstructed image, denoted as $Res(x)$. Finally, $Res(x)$ is used to directly input ClsNet for distinguishing authenticity. 
It is noted that the reconstruction model and the classification model are trained simultaneously.
In the training process, the suitable dynamic balance point between multiple tasks is trained by minimizing the equation (1).

\begin{figure*}[ht]
    \centering
    \includegraphics[width=0.91\textwidth]{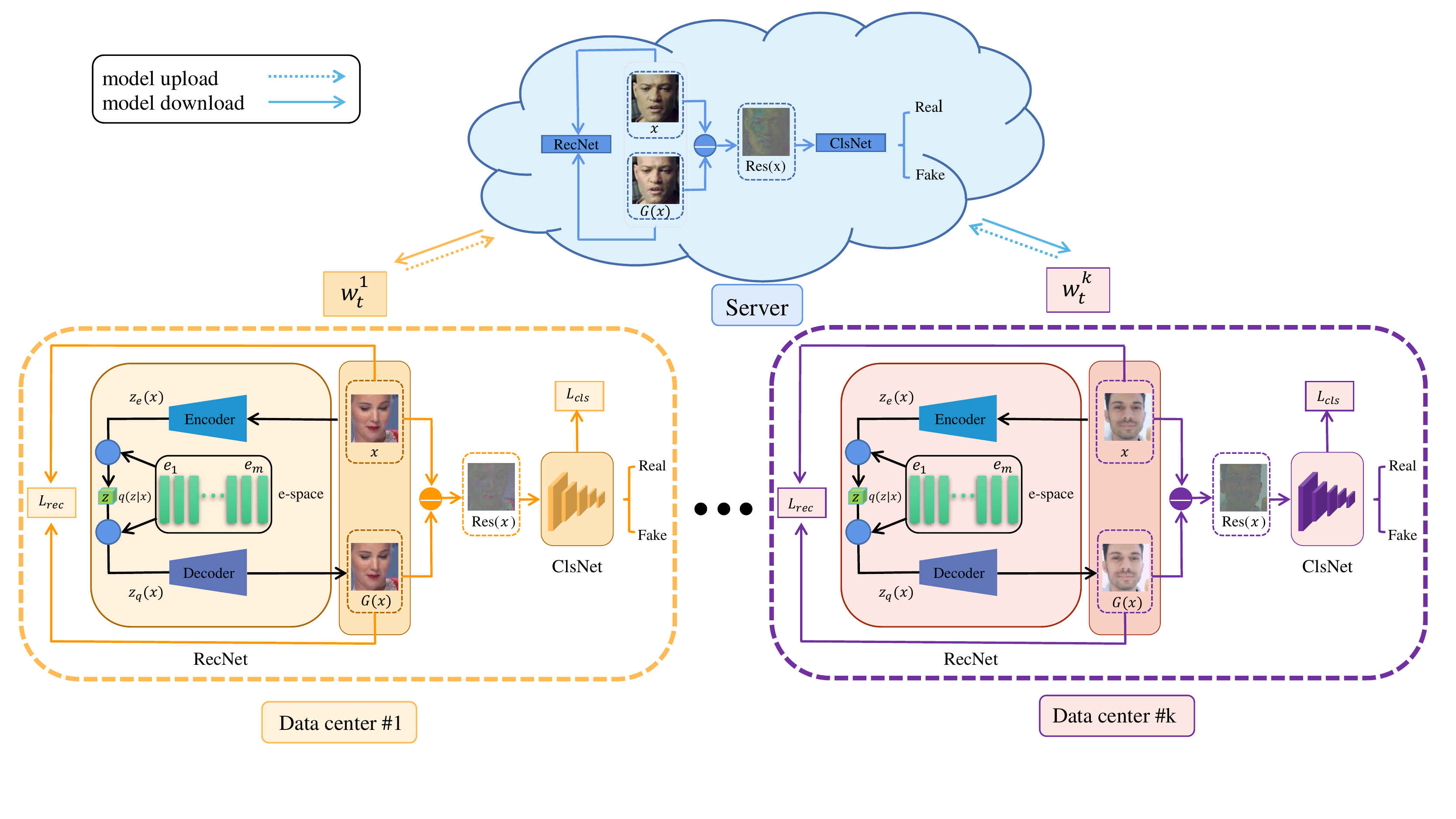}
    \caption{The framework of a novel generalized residual Federated learning for face Forgery detection (FedForgery). Each data center uses its own data to train the model locally, and the local model further analyzes the difference between real image and artifact reconstruction residuals under different artifact types through a variational autoencoder. After a round of local training is completed, each data center uploads the obtained model parameters to the global server for weighted aggregation. After the aggregation is completed, the global server sends the updated parameters to each data center. This communication process continues until the model converges and is eventually tested using the updated aggregated global server model.}
\end{figure*}
\label{figure_framework}
\noindent

We jointly train parameters in mentioned RecNet and ClsNet by optimizing the proposed objective, which combines the training objective $L_G$ of VQ-VAE\cite{van2017neural}, the pixel reconstruction loss $L_{rec}$ and the forgery classification loss $L_{cls}$.

\begin{equation}
L=\mu_{1}L_{G}+\mu_{2}L_{rec}+\mu_{3}L_{cls}.
\end{equation}

Here the variational autoencoder $G(\cdot )$ aims to generate input face with high quality.
Inspired by the good performance of the discrete latent model in image generation tasks, we also introduce the discrete representation learning to construct $G$.
The RecNet module consists of three components: the encoder, the decoder and the potential embedding space.
The embedding space $e\in R^{m \times d}$ is defined here, $m$ is the size of the discrete embedding space, $d$ is the dimension of each embedding vector in this space.
The encoder is denoted as $E_{\varphi}(x)$, which is parameterized by $\varphi$.
{Parameters   and $\mu_1$, $\mu_2$ and $\mu_3$ indicate the weights of the generative model, pixel reconstruction and classification terms. The higher quality of the reconstructed image may bring adverse effects in detection forgery clues, which focus on real data distribution but ignore the discrepancy. Thus, both hyperparameters $\mu_2$  and $\mu_3$ balance the importance of the influence of forgery detection discriminability.
Thus, the training objective function is followed:

\begin{equation}
L_G=\alpha||sg[E_{\varphi}(x)]-e||_{2}^{2}+\beta||E_{\varphi}(x)-sg[e]||_{2}^{2},
\end{equation}
where the former term is the alignment loss, which aims to make the embedding vectors $e$$_{i}$ closer together the encoder outputs $E$$_\varphi$($x$) as possible.
sg[·] stands for stop gradient.
The stop gradient operator on the encoder output indicates that the term is only used to update the embedding vector, which is conducive to better learning the embedding space. 
The latter term aims to force the encoder output to commit as much as possible to its closest embedding vector. 
The stop gradient on the embedding vector helps to limit the volume of the discrete embedding space and prevent its free growth.
Parameters $\alpha$ and $\beta$  indicate the weights of the embedding and encoder alignment terms, which also balance the importance of generative model performance.
The settings of hyper-parameters $\alpha$ and $\beta$ are shown in the following.

To  reduce the difference between the reconstructed face and the input face, formula (3) is introduced here to optimize the parameters of the encoder and decoder:

\begin{equation}
L_{rec}=||x-G(x)||_{2}^{2},
\end{equation}
where $x$ is the input face, and $G(x)$ is the constructed images with the variational autoencoder $G(\cdot )$.

Since face forgery detection is essentially considered as a binary classification task, the cross-entropy loss function $L$$_{cls}$ is used in formula (4) to evaluate the forgery detection performance as followed:

\begin{equation}
L_{cls}=-log\ P(Res(x),y),
\end{equation}
where $P$ means the prediction probability of the residual $Res(x)$ through the ClsNet.
$y$ is the label of the input face $x$. 
In the following experiments, hyper-parameters are used to control the trade-off between reconstruction and classification tasks. 
According to the experiment analysis shown in Section IV, we set the hyperparameters to be $\mu_1$=1, $\mu_2$=1, $\mu_3$=1, $\alpha$=1, $\beta$=4 respectively.

\subsection{Residual Federated Learning}
To further improve the representation discriminability to detect different forgery artifact patterns, we design the residual federated learning strategy. The designed global model not only updates parameters through training decentralized data collaboratively to protect privacy, but also aggregates the captured discrepancy information in several local models in a federated learning strategy.
As shown in Figure 2, the residual federated learning framework is designed for the distributed forgery detection models deployment, which aims to mimic real-world scenarios. 
There are $K$ data centers in diverse local clients and forgery face datasets $D$ are distributed and stored in each data center ($D$$_{1}$...$D$$_{K}$), satisfying $D$={$D$$_{1} \cup...\cup$$D$$_{K}$} and $D$$_{1}\cap...\cap$$D$$_{K}$=$\varnothing$.  
Additionally, the data set between each data center should be independent and cannot be exchanged. 
Here each data center has a local face forgery detection model ($FFD$).
The global server will update model parameters according to these local parameters uploaded by each data center. The local $FFD$ model  by the $k$-th data center is parameterized by $\omega$ $^{k}$ ($k$ = 1, 2, 3, ... ,  $K$), denoted as $FFD(\omega^{k})$. 
As shown in formula (5), $L$ is the loss of the forgery detection model in each data center, its specific form is in equation (1).
 
\begin{equation}
L_{FFD(\omega^{k})}=\sum_{(x,y)\sim D_{k}}L(x,y).
\end{equation}

The global server will calculate aggregated model parameters according to formula (6) as followed:
\begin{equation}
\omega\leftarrow \sum_{k=1}^K p_k\omega^{k},
\end{equation}
where p$_{k}$ is the proportion of the number of data points of the $k$-th participant to the total number of data points, which is also the weight of each local model when aggregated.
 
To improve the forgery detection accuracy, we further extend the number of communication rounds between the global server and the data center to $t$ rounds.
The detailed update strategy is shown in formula (7) as followed:
 \begin{equation}
\omega_{t}\leftarrow \sum_{k=1}^K p_k\omega_{t}^k,
\end{equation}
where $\omega_t^k$ represents the model parameters owned by the $k$-th data center in the $t$-th round of update. The initial parameters of the next round of data center models are issued after the weighted average of the global server.  
After $t$ rounds updates, the global server model can be obtained without accessing data privacy in local clients. 

\begin{table}[!ht]
\begin{tabular}{p{0.95\columnwidth}}

\hline
\textbf{Algorithm 1.} FedForgery\\
\hline
\textbf{Require}: Local data center have $K$ Data centers  $D$$_{1}$, ... , $D$$_{K}$;\\
\textbf{1}: $K$ data centers have $K$ face forgery detection models($FFD$) parameterized by $w^{K}$, $E$ is the number of epochs for the local train, $\lambda$ is the learning rate, $t$ is the number of rounds of global communications between the local data center and global server;\\
\textbf{2}: Global Server Aggregates: initialize $w_{0}$\\
\quad \quad for each round $t$=0, 1, ... , do \par
\quad \quad \quad for each Data Center $k$=1, 2, ... , $K$ in parallel do\\
\quad \quad \quad $\omega_t^k$ $\leftarrow$ Data Center Update($k$, $\omega_t$)\\
\quad \quad \quad end for\\
\quad \quad \quad $\omega_{t}$ $\leftarrow$ $\sum_{k=1}^K$ $p_{k}$ $\omega_{t}^k$\\
\quad \quad \quad Download $\omega_t$ to Data Centers\\
\quad \quad end for\\
\textbf{3}: Data Center Update($k$, $\omega$)\\
\quad \quad for each local epoch $i$ from 1 to $E$ do\\
\quad \quad \quad $min\ L_{FFD(\omega^{k})}=\sum_{(x,y)\sim D_{k}}L(x,y)$\\
\quad \quad  \quad $\omega^{k}$ $\leftarrow$ $\omega^{k}$ $-$ $\lambda$ $\nabla$ FFD($\omega^{k}$)\\
\quad \quad end for\\
\quad \quad Upload $\omega^k$ to global server\\
\hline
\end{tabular}
\vspace{-3mm}
\label{Algorithm}
\end{table}

The detailed procedures of the proposed FedForgery are shown in \textbf {Algorithm 1}.
The variational autoencoder analyzes the difference of the reconstruction residual $Res(x)$ between real and fake images.
The classification network ClsNet is used to process the residual image $Res(x)$ to output the final prediction result. 
The federated learning distributed framework is used to train and deploy the forgery detection model, and upload model parameters $\omega$ from different local data centers to the global server for boosting generalization and protecting data privacy.
Noting that the proposed FedForgery not only protects the data privacy of non-public videos through distributed storage, but also improves the discriminative representation avoid overfitting to distinguish the authenticity.


\section{Experiments}
In this section, we first introduce the representative public face forgery datasets: FaceForensics++, WildDeepfake and Deeperforensics-1.0.
Then the self-constructed datasets Hybrid-domin forgery dataset and generalized forgery dataset are introduced to evaluate the performance in hybrid dataset and generalized dataset face forgery detection tasks.
In addition, we further evaluate the novel generalized residual federated learning for face Forgery detection algorithm and compare it with state-of-the-art methods. 

\subsection{Datasets}
\label{Datasets}
In this paper, we use three public datasets: FaceForensics++ dataset (abbreviated as FF++), WildDeepfake dataset and Deepforensic-1.0 dataset.
Example samples are shown in Figure 3. 
It is noted that the large-scale Deepforensic-1.0 dataset can help evaluate the method's ability for real-world face forgery detection tasks.
In addition,  we constructed two datasets: Hybrid-domain forgery dataset and generalized forgery dataset to mimic complex real-world scenarios.

\textbf{FaceForensics++} The FaceForensics++ dataset \cite{rossler2019faceforensics++} is a benchmark dataset for evaluating face forgery detection methods. It consists of 1000 real videos extracted from YouTube, in addition to using four different algorithms to generate their corresponding fake videos, respectively DeepFakes (DF), Face2Face (F2F) \cite{thies2016face2face}, FaceSwap (FS) and NeuralTextures (NT) \cite{thies2019deferred}. In addition, there are three different compression factors, corresponding to the original video (c0), the high compression rate video (c40), and the lower compression rate video (c23). Our experiments are all based on the c23 compression format.

\textbf{WildDeepfake}
The WildDeepfake dataset contains 7314 face sequences extracted from 707 deepfake videos \cite{zi2020wilddeepfake}. The videos in WildDeepfake are all collected from the Internet, and the sources are different, and some videos may have been compressed many times, which makes the detection of the WildDeepfake dataset more challenging, so it is often used as a benchmark video to evaluate the performance of deepfake detection models. 

\textbf{Hybrid-domain forgery dataset}
To effectively evaluate the proposed FedForgery performance in mentioned hybrid-dataset face forgery detection task, we utilize these two public datasets to construct a novel Hybrid-domain forgery dataset.
The details of the designed protocol are shown as follows: combine four diverse forgery subtypes of the FF++ dataset and the WildDeepfake dataset into the whole dataset with five different artifact types;
the training set contains 100,000 images where true images have the same number of fake images;
the ratio of the training set and testing set is kept at 7: 3. 

\textbf{Generalized forgery dataset}
To effectively evaluate the proposed FedForgery performance in mentioned generalized-dataset face forgery detection task, we utilize these two public datasets to construct a novel generalized forgery dataset.
The details of the designed protocol are shown as follows: 
the forgery artifact types in the training set of each data center are diverse, and the forgery artifact type in the testing set of the global server is different from the training set;
choose four artifact types of the training set from the mentioned Hybrid-domain forgery dataset each time, and the last artifact type is selected as the testing set on the global server.

\textbf{Large-scale Deeperforensics-1.0} The Deepforensics-1.0 dataset\cite{jiang2020deeperforensics}  is a large-scale and widely used real-world face forgery detection benchmark, which consists of 1000 high-quality videos in total. The fake videos are generated by an end-to-end face swap framework. The split ratio of the training, validation and test set is 7:1:2. With the same protocol in \cite{jiang2020deeperforensics}, we compare the proposed FedForgery with representative algorithms to prove the strong avoid overfitting.

 \begin{figure}[ht]
 \centering
    \includegraphics[width=0.48\textwidth]{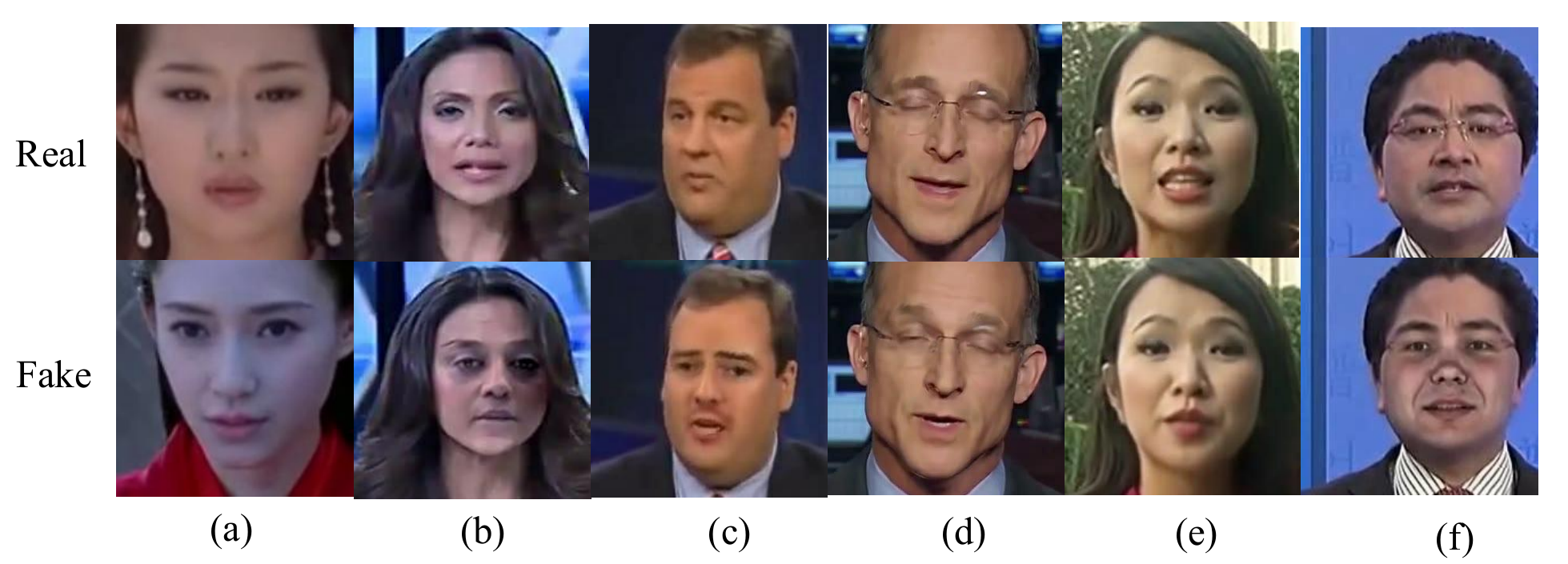}
\caption{The samples of public face forgery datasets: (a) WildDeepfake. (b) DeepFakes. (c) Face2Face. (d) FaceSwap. (e) NeuralTexture. (f) Deepforensics-1.0.}
\end{figure}

\subsection{Implement Details}
\label{Implement Details}

\textbf{Preprocessing} 
We select the Dlib \cite{jeong2021eye} tool to detect faces to reduce disturbance of complex background.
Next, we normalize all aligned faces and resize them to the size of 296×296.
We perform frame-by-frame extraction on the video dataset.
The proposed deep network model is implemented on the PyTorch \cite{paszke2019pytorch} platform. 
We utilize a variational autoencoder with several convolutional layers \cite{walker2016uncertain} to reconstruct input images and the pre-trained Resnet 50 \cite{al2020breast} as the classifier following. 
We employ SGD with a momentum factor of 0.5. The initial learning rate is 0.01.

\textbf{Federated learning settings} 
In the following experiment, we set 10 diverse data centers as local clients, and allocate the same number of datasets to each data center. 
The local training batch size is set to 32. 
All data centers participate in the aggregation update process of model parameters.
After a round of completing the training steps in each data center, we upload local center training model parameters to the global server to aggregate and update the parameters.
The global server then sends the updated parameters to each local data center. 
Finally, the global server model is evaluated on the testing set.

We conducted experiments on the settings of cyber-parameters in the FedFogery method. 
For the convenience of experiments, we sampled 10\% images from the WildDeepfake dataset. 
Here we utilized the simple grid search strategy to choose optimal hyper-parameters manually. 
To promise the choices of hyperparameter choice is optimal for other kinds of cases, we conduct the optimal parameter choices in two different kinds of datasets (WildDeepfake and Hybrid-domain forgery datasets) as shown in Figure 4.

\textbf{Influence of parameter $\beta$}  
To evaluate the effect of parameter $\beta$ on the evaluation performance, parameters of different sizes are selected to conduct experiments on different kinds of datasets. 
In the left subgraph of Figure 4, we evaluate the effect of parameter $\beta$  from a set of \{0.10, 0.50, 1.00, 2.00, 3.00, 4.00, 5.00, 7.00\} is illustrated when we set parameter $\mu_2$ as 1 and $\mu_3$ as 1. We find that when the parameter $\beta$ is set as 4, the forgery detection accuracy becomes better. It is because the suitable weight of encoder alignment terms could improve generative model performance.

\textbf{Influence of parameter $\mu_2$}  
We also investigated the effect of another parameter $\mu_2$ on different kinds of datasets. 
In the middle subgraph of Figure 4, we evaluate the effect of parameter $\mu_2$ from a set of \{0.10, 0.30, 0.70, 1.00, 2.00\} is illustrated when we set parameter $\beta$ as 4 and $\mu_3$ as 1. The weights of the pixel reconstruction would affect the exploration of forgery clues in the pixel level. We find that when parameter $\mu_2$ reaches approximately 1, the detection performance improves.

\textbf{Influence of parameter $\mu_3$}  
We evaluate the effect of parameter $\mu_3$ according to the evaluation performance on different kinds of datasets. 
In the right subgraph of Figure 4, we test the effect of parameter $\mu_3$ from a set of \{0.01, 0.10, 0.50, 1.00, 5.00, 10.00\} is illustrated when we set parameter $\beta$ as 4 and $\mu_2$ as 1. The weights of the classification loss terms are the key parameters to improve the representation discriminability. However, higher values of parameter $\mu_3$ may cause overfitting in the specific data domain. We find that when parameter  $\mu_3$ achieves approximately 1, the performance achieves the best.
The parameter $\alpha$, $\mu_1$ are set as the fixed value of 1 with our experience.
Experimental results prove these chosen hyperparameters are optimal and applicable in most similar kinds of cases.

 \begin{figure}[ht]
 \centering
    \includegraphics[width=0.48\textwidth]{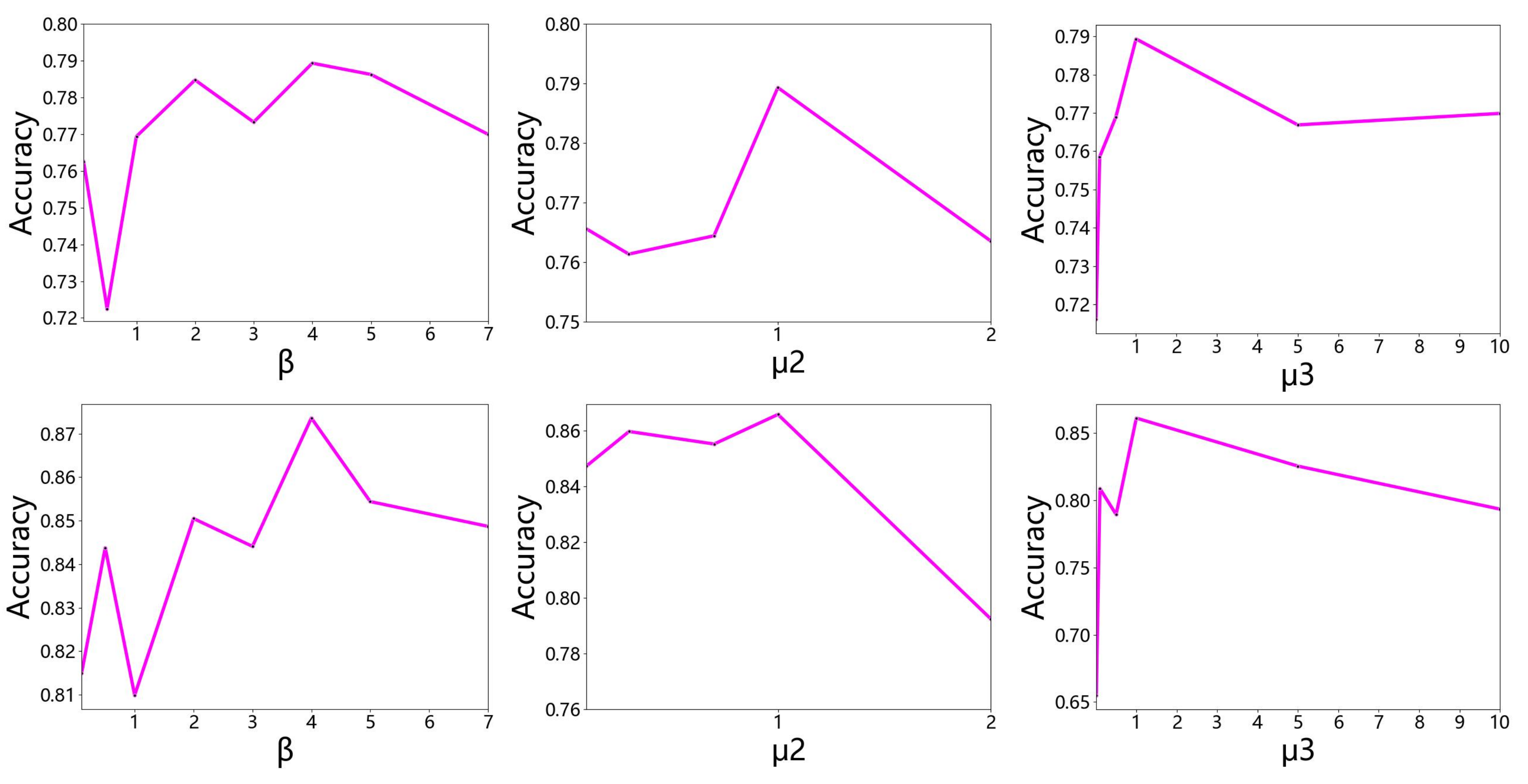}
\caption{In the first row of subplots of this figure, the left subfigure shows the evaluation accuracies of different numbers of the parameter $\beta$ on WildDeepfake; The middle subfigure shows the evaluation accuracies of different numbers of the parameter $\mu_2$ on WildDeepfake; The right subfigure presents the evaluation accuracies of different numbers of the parameter $\mu_3$ on WildDeepfake. In the second row of subplots of this figure, the left subfigure shows the evaluation accuracies of different numbers of the parameter $\beta$ on Hybrid-domain forgery dataset; The middle subfigure shows the evaluation accuracies of different numbers of the parameter $\mu_2$ on Hybrid-domain forgery dataset; The right subfigure presents the evaluation accuracies of different numbers of the parameter $\mu_3$ on Hybrid-domain forgery dataset.}
\end{figure}



\subsection{Comparison Results}
\label{Comparison Results}
In this section, we compare our proposed method with state-of-the-art methods on the large-scale Deeperforensics-1.0 face forgery datasets, Hybrid-domain forgery dataset and the Generalized forgery dataset separately. To sufficiently evaluate forgery detection performance, we utilized the classification accuracy and the area under the receiver operating characteristic curve (AUC) as the quantitative metrics.

\begin{table}[htbp]
	\centering
	\caption{Evaluation accuracy rate (in \%) and area under the receiver operating characteristic curve (in \%) of forgery detection performance on constructed hybrid-domain forgery dataset by several state-of-the-art methods. The best result is displayed in black font and the second place is underlined.}
	\label{tab:1}  
	\begin{tabular}{c|cc}
		\hline

		 Methods & Accuracy(\%) & AUC(\%)
		\\
		\hline
 ~&
\multicolumn{2}{c}{\textbf{Without Considering Privacy Issue}} \\
\hline
		KNN \cite{dzanic2020fourier} & 53.23 &55.01 \\
		CNNDetection$_1$ \cite{wang2020cnn} & 74.73  & 83.36 \\
		Xception \cite{chollet2017xception} & 76.31 & 86.06 \\  
      	CNNDetection$_2$ \cite{wang2020cnn} & 78.21  & 88.05 \\
	    RECCE \cite{cao2022end} & 85.03 & 92.75\\
     GFFD \cite{luo2021generalizing} & 86.56 & 93.21\\
    RFM \cite{wang2021representative} & 83.85 & 92.73\\
     DCL \cite{sun2022dual} & \underline{86.82} & $\textbf{94.48}$\\
    FedForgery* (Ours) & $\textbf{87.36}$ &\underline{93.23}\\
    
\hline
~&
\multicolumn{2}{c}{\textbf{Considering Privacy Issue}} \\
\hline
   FedForgery (Ours) & 85.55 & 91.12\\
	\hline
	\end{tabular}
\end{table}
\begin{table*}[htbp]
	\centering
	\caption{Evaluation accuracy rate (in \%) and area under the receiver operating characteristic curve (in \%) of forgery detection performance on constructed generalized-domain forgery dataset by several state-of-the-art methods. The best result is displayed in black font and the second place is underlined.}
	\label{tab:2}  
	\begin{tabular}{c|ccccc}
		\hline	
		Methods &  WildDeepfake & DeepFakes& Face2Face & FaceSwap & NeuralTextures		\\
		\hline
  ~&
  \multicolumn{5}{c}{\textbf{Without Considering Privacy Issue}} \\
		\hline
	    CNNDetection$_1$ \cite{wang2020cnn} &50.44$\backslash$58.34&58.68$\backslash$66.54&49.64$\backslash$52.43&55.30$\backslash$61.22&51.07$\backslash$53.99   \\
		KNN \cite{dzanic2020fourier}&51.10$\backslash$51.55&50.57$\backslash$51.03&51.31$\backslash$52.02&52.18$\backslash$52.99&50.70$\backslash$51.01\\
		Xception \cite{chollet2017xception}&60.83$\backslash$66.81&65.18$\backslash$72.14&46.54$\backslash$50.57&55.95$\backslash$62.60&50.05$\backslash$55.12 \\
      	CNNDetection$_2$\cite{wang2020cnn} & 62.54$\backslash$70.86&63.19$\backslash$69.44&51.53$\backslash$57.36&59.59$\backslash$71.05&54.02$\backslash$58.59 \\
	    RECCE\cite{cao2022end}&65.70$\backslash$68.75&69.54$\backslash$77.08&53.93$\backslash$57.01&68.46$\backslash$71.69&59.39$\backslash$70.83\\
	 GFFD \cite{luo2021generalizing}&64.42$\backslash$\textbf{80.93}&63.84$\backslash$63.89&52.30$\backslash$64.75&68.63$\backslash$\underline{83.30}&57.75$\backslash$\underline{77.17}\\ 
   RFM \cite{wang2021representative}&59.98$\backslash$73.79&67.51$\backslash$75.12&\textbf{60.98}$\backslash$67.51&70.23$\backslash$76.39&59.55$\backslash$64.40\\ 
   		DCL \cite{sun2022dual}&64.20$\backslash$75.40&71.14$\backslash$\underline{81.93}&52.91$\backslash$61.19&67.83$\backslash$79.46&55.33$\backslash$70.77\\
FedForgery* (Ours) &\underline{67.75}$\backslash$\underline{75.45}&\textbf{73.47}$\backslash$\textbf{82.08}& 53.77$\backslash$\underline{68.41}&\textbf{71.13}$\backslash$82.21&\textbf{60.95}$\backslash$\textbf{80.32}\\
\hline
~&
\multicolumn{5}{c}{\textbf{Considering Privacy Issue}} \\
\hline
FedForgery (Ours) &\textbf{68.03}$\backslash$74.59&\underline{72.24}$\backslash$79.26& \underline{54.74}$\backslash$\textbf{69.66}&\underline{70.36}$\backslash$\textbf{83.79}&\underline{60.58}$\backslash$76.64\\ 
	\hline
	\end{tabular}
\end{table*}
\textbf{Results on Hybrid-domain forgery dataset}
As shown in Table I, we reimplemented several representative face forgery methods on the Hybrid-domain forgery dataset. 
CNNDetection \cite{wang2020cnn} aims to design a simple convolutional neural network to detect forgery clues.
Related works\cite{luo2021generalizing} found that CNN detectors are prone to overfitting the texture patterns of specific generation methods.
Thus, two different data augmentation methods were designed to boost performance.
KNN \cite{dzanic2020fourier} uses the frequency domain information of pictures to detect forgery. This method may be misled by some noise signals, especially when encountering complex datasets, and its detection performance will be greatly affected.
RECCE \cite{cao2022end} enables the classifier to learn a more general representation by learning from real images, but its defect is that it cannot protect the privacy of training data in the face forgery detection task.
 GFFD \cite{luo2021generalizing} utilized high-frequency noise and the correlation between complementary modalities to facilitate feature learning. RFM \cite{wang2021representative} explored sensitive local regions and specific data augmentation strategies to boost performance. DCL \cite{sun2022dual} designed the inter-instance contrastive learning and local content inconsistencies to detect clues. 
 But the defect is that it cannot protect the privacy of training data in the face forgery detection task.

  Noting that the proposed FedForgery belongs to distributed training method, enables sensitive personal information to be stored on the local clients, different from these traditional centralized training methods without considering privacy issues. 
  Related research \cite{zhang2021survey, Fed_ReID} states that the efficiency and accuracy of advanced federated learning models are getting closer and closer to centralized training models, but still slightly inferior to that of centralized training. For a fair comparison, we add the proposed methods with the centralized training strategy, denoted as FedForgery* in the following experiments. 
  As shown in Table I, benefiting from the designed generalized contractive feature learning, DCL \cite{sun2022dual} achieves the best AUC performance. However, our proposed method is slightly below 1.25\% at AUC performance, but achieves better accuracy at 87.36\% compared with DCL \cite{sun2022dual}, GFFD \cite{luo2021generalizing} and RFM \cite{wang2021representative}. It is because the designed robust residual features learning could effectively capture domain-invariant discrepancy information between real and fake images. 
 
\textbf{Results on Generalized forgery dataset}
To further verify the generalization of the proposed FedForgery, we conduct new experiments on the Generalized forgery dataset. 
The aim is to distinguish the authenticity of input faces even with unknown artifact types.
Similarly, we reimplement representative face forgery detection methods for fair comparisons.
The experimental results are shown in Table II.
For convenience, we set the testing set only contains one artifact type, and the training set contains the rest four types.
For example, when the testing set is chosen from DeepFakes dataset, the training set contains artifact types from WildDeepfake, Face2Face, FaceSwap, NeuralTextures datasets.


As shown in Table II, we construct the generalized forgery dataset to evaluate the ability of methods to explore forgery clues in unknown patterns. It is obvious that our proposed method achieves better performance than DCL \cite{sun2022dual} in all scenarios, and also better than GFFD \cite{luo2021generalizing} and RFM \cite{wang2021representative} in most scenarios with accuracy and AUC metrics. 
It is encouraging that when considering privacy issue, our method FedForgery not only maintain similar performance with centralized training FedForgery*, but also achieve better performance in WildDeepface, Face2Face and FaceSwap testing sets. 
\emph{It proves the proposed residual federated learning strategy could aggregate diverse discrepancy information in local clients and help learn more essential discrepancies to boost the generalization ability. 
}
We believe it will inspire researchers to explore distributed learning and generalization analysis in the community.

\begin{table}[htbp]
	\centering
	\caption{Evaluation accuracy rate (in \%) of forgery detection performance on the Deeperforensics-1.0 dataset by several state-of-the-art technologies. The best result is displayed in black font and the second place is underlined.}
	\label{tab:3}  
	\begin{tabular}{c|cc}
		\hline	
		Methods  & \makecell{std\\  std} & \makecell{std\\ std/sing}
		\\
		\hline
 ~&
\multicolumn{2}{c}{\textbf{Without Considering Privacy Issue}} \\
\hline

		C3D \cite{tran2015learning}& 98.50 & 87.63 \\
		TSN \cite{wang2016temporal} & 99.25  & \underline{91.50} \\
		I3D \cite{carreira2017quo} & \textbf{100.00} & 90.75  \\ Resnet+LSTM\cite{he2016deep} \cite{hochreiter1997long} & \textbf{100.00}  & 90.63 \\
	   XceptionNet \cite{chollet2017xception} & \textbf{100.00} & 88.38\\
            \hline
~&
\multicolumn{2}{c}{\textbf{Considering Privacy Issue}} \\
\hline
    FedForgery (Ours) & \underline{99.75} & \textbf{95.21}\\
	\hline
	\end{tabular}
\end{table}

\textbf{Results on large-scale Deepforensics-1.0 dataset}
As shown in Table III, we compare our proposed algorithm with other representative face forgery detection methods with a similar protocol in \cite{jiang2020deeperforensics}. To prove the strong generalization ability, we evaluate comparison methods with dataset perturbations. Here we select 1000 manipulated videos in the standard set (std), and 1000 manipulated videos with single-level distortions (std/sing). 
When the training set is the standard set and the testing set is also the standard set, all these comparison methods could achieve high accuracies ($\textgreater 98\%$), which may benefit from these similar distributions. 
The proposed FedForgery not only considers the privacy issue through distributed training, but also achieves accuracy at 99.75\%, which is already close to the ultimate accuracy. Thus, the first protocol is not a reasonable experimental setting to distinguish the pros and cons of these comparison methods. 
To further evaluate the generalization ability, we utilized the dataset perturbations to mimic real-world scenarios. When the training set is the standard set and the testing set is with single-level distortions (std/sing), the proposed FedForgery can achieve the SOTA accuracy and even exceed Resnet+LSTM \cite{he2016deep}\cite{hochreiter1997long} by 4.58\%, XceptionNet \cite{chollet2017xception} by 6.83\% and I3D\cite{carreira2017quo} by 4.46\%. We think these accuracy improvements in the large-scale dataset are benefitted from the proposed residual federated learning that could boost the representation generalization ability.

\section{Discussion}

\subsection{Ablation study}
To verify the effectiveness residual federated learning framework for face forgery detection, we conducted the following experiments as shown in Figure 5. 
For the convenience of analysis, we sampled 10\% images from the WildDeepfake Dataset \cite{zi2020wilddeepfake} in the following experiments. 
We set eight diverse data centers as local clients, and allocate the same number of datasets to each data center. The eight data centers would evaluate forgery detection performance after the local client training.
And the global server could evaluate performance after each round of global communication. 
As shown in Figure 5, the pink line indicates the detection performance of the global server model, and the blue line indicates the average detection accuracies of eight data centers. 
Here we find the performance of the global server is significantly higher than the results of data centers in local clients.
The forgery accuracy improves from about 1.34\% to 6.35\% after applying the proposed residual federated learning strategy.
The curve clearly proves that the proposed FedForgery can improve the forgery detection performance by aggregating the model parameters, and meanwhile protect the privacy of training data in local data centers. 
Within a certain number of communication times, with the increase in global communication times, the detection performance of the proposed FedFogery is also improved. 
When the global communication times reached about ten times, the forgery detection performance of the global server reaches the best point, with an accuracy rate of 79.21\%. It proves the proposed residual federated learning strategy could help the forgery detection model not only train weights collaboratively for data security, but also would maintain strong performance in forgery detection tasks.

  \begin{figure}[ht]
    \centering
    \includegraphics[height=4.3cm, width=7cm]{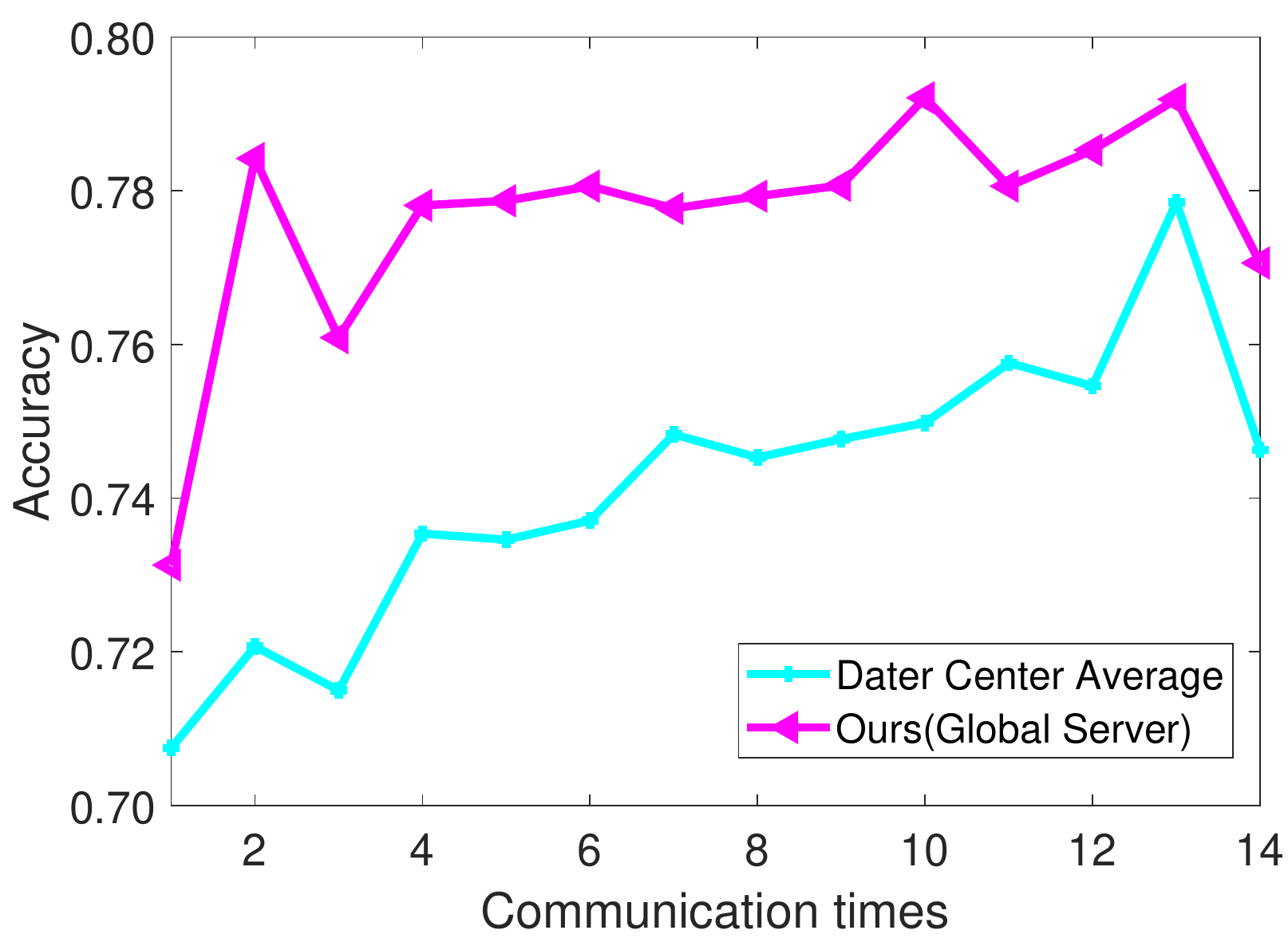}
    \caption{The face forgery detection evaluation accuracies of  federated learning framework on WildDeepfake dataset.}
\end{figure}



To further evaluate the performance of the residual feature learning strategy of the proposed algorithm, we design the ablation study experiments on Hybrid-domain face forgery dataset. For the convenience of comparison, the same parameters setting is utilized as elaborated in Section IV-B. The accuracy and AUC separately degrade at 83.36\% and 91.02\% when directly removing the residual feature instead of the raw images. It is because exploring the specific characteristics of reconstruction residual is easier than inputting raw images in the multiple domain forgery scenarios. The proposed residual feature learning strategy could avoid overfitting and effectively capture domain-invariant discrepancy information between real and fake images, which decreases the accuracy by 2.19\% on the Hybrid-domain face forgery dataset.
\subsection{Analysis of residual feature}
The designed discriminative residual feature learning goals to extract discriminative forgery clues even for unknown artifact types. Existing traditional forgery detection methods generally directly input the raw image into an end-to-end network for binary classification. However, when processing data from complex unknown artifact patterns, the forgery detection performance is not encouraging. Different from direct feeding raw images into the network, we utilized a variational autoencoder framework to reconstruct the image and extract the residual feature to explore forgery clues.
In order to verify the effectiveness of the proposed residual learning strategy, we utilize the t-SNE \cite{van2008visualizing} to visualize the feature distribution when inputting original images and residual images separately in the  data center.
For the fairness of the comparison, we extract features in the layer before the fully connected layer in the ResNet50 network for original images.
The results are shown in Figure 6.

We select 1500 authentic images and 1500 fake images from the Hybrid-domain forgery dataset here. 
It can be observed that the feature distribution of the fake images represented by the green part (in the left subfigure of Figure 6) is scattered and overlap with the red part. 
It is because the classifier has a poorer ability to discriminate the forged type when inputting original images.
As shown in the right subfigure of Figure 6, the feature distribution of fake images represented by the green part is more concentrated, indicating that fake images are embedded into a relatively compact feature space. This also demonstrates that our proposed method could capture more discriminative common representations for forgery detection. 
Thus, the proposed residual learning strategy can effectively improve the generalization ability of forgery detection.

\begin{figure}[!t]
\centering
\subfloat{\includegraphics[scale=0.32]{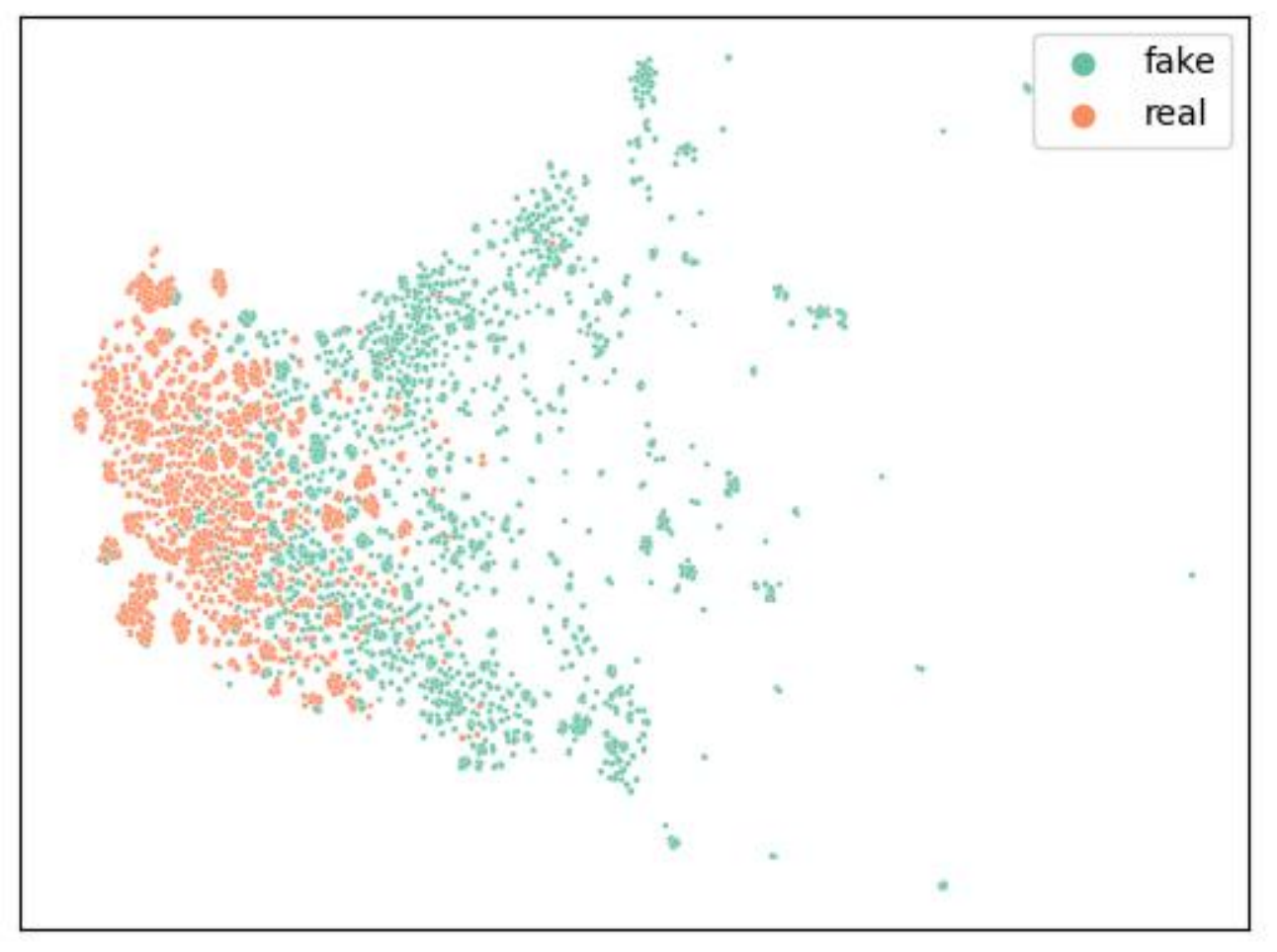}%
\label{fig_first_tsne}}
\hfil
\subfloat{\includegraphics[scale=0.32]{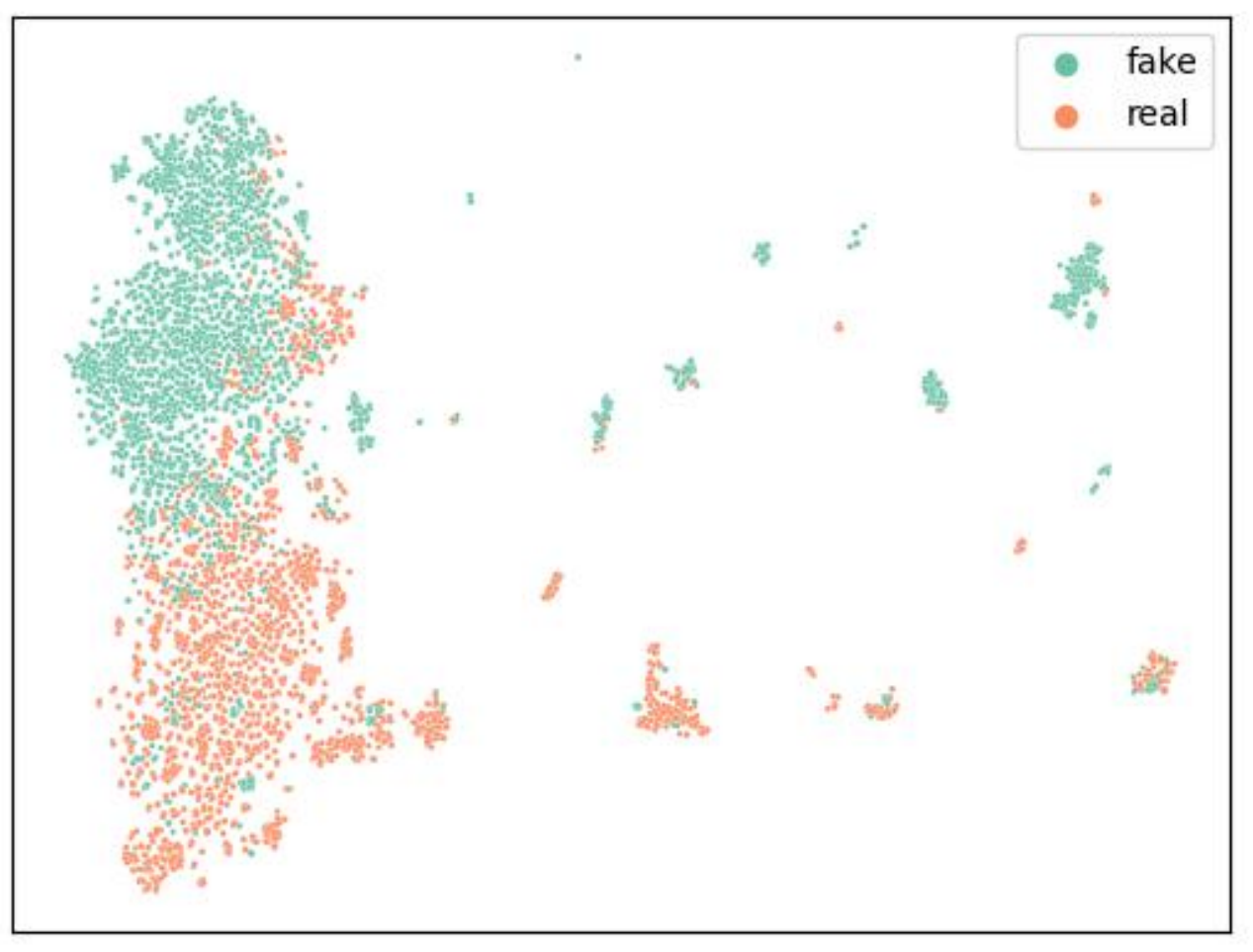}%
\label{fig_second_tsne}}
\caption{The left subfigure shows the feature distribution learned by the data center directly using ResNet50 on the original images; The right subfigure presents the feature distribution of our method on the reconstructed residual images.}
\label{fig_tsne}
\end{figure}

\subsection{Failure cases}

As shown in Figure 7, we utilized the Grad-CAM algorithm as the visualization tool to analyze failure cases.
For example, the left two samples show that the detection model pays much attention to the margin region, which would make wrong judgments. 
The right two samples show the detection model is affected by the complex background.
Thus, face forgery detection in the wild is still a challenging problem.

\begin{figure}[ht]
 \centering
    \includegraphics[width=0.30\textwidth]{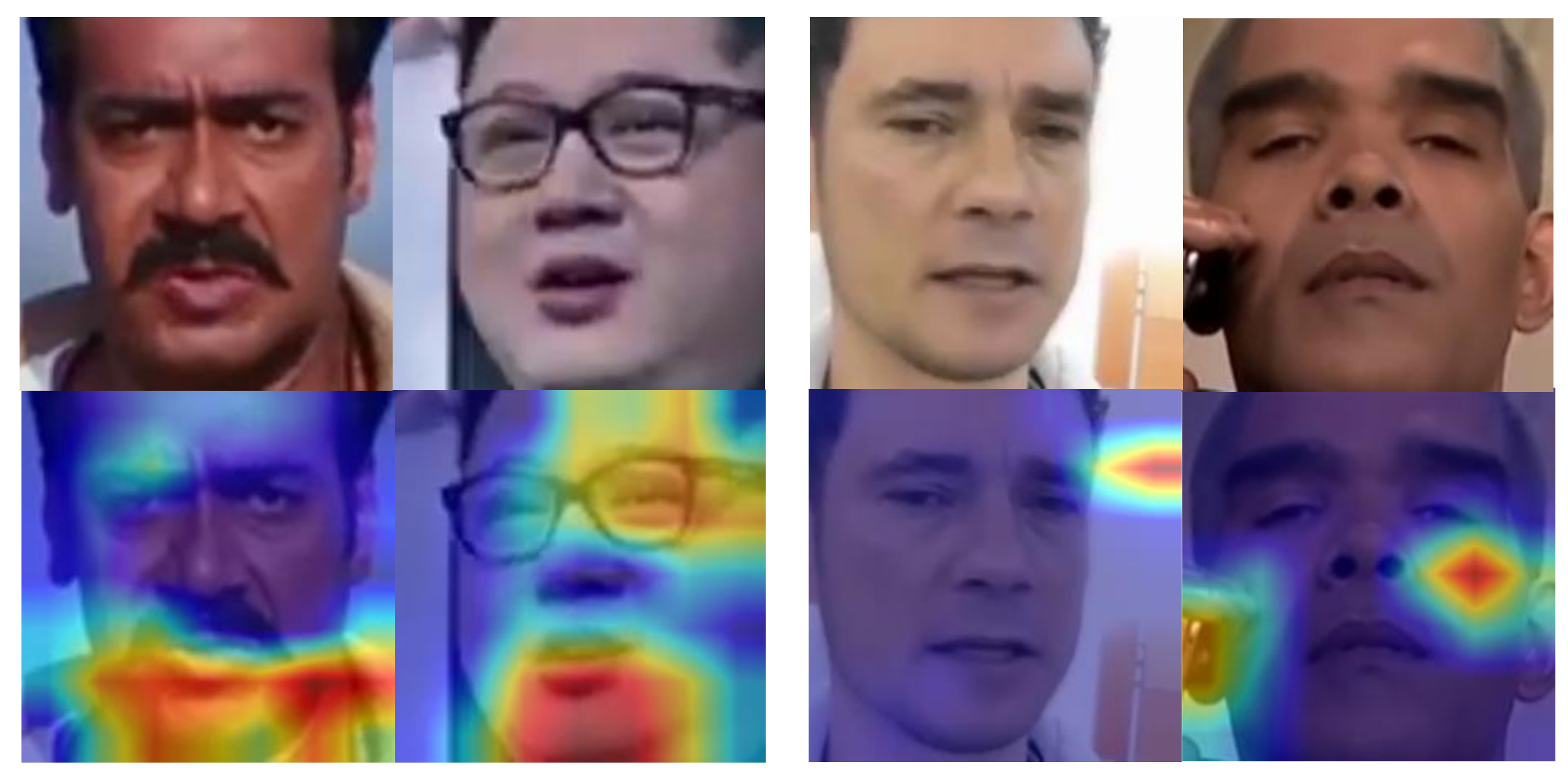}
\caption{Grad-CAM activation map for failure cases on WildDeepfake dataset.}
\end{figure}

\subsection{Visualization Results}

As shown in Figure 8, in order to analyze the visualization results of each data center and global server focusing on different regions, we utilized the gradient-based visualization tool to generate a Grad-CAM based attention map \cite{selvaraju2017grad}.  It reveals the differences in the range of regions that data centers and the global server rely on for forgery detection.
 As shown in subfigures (a)-(d), each data center and global server network focuses on the local facial semantic regions, which would be helpful for authenticity identification. 
 
In Figure 8 (a), the first eight columns represent the visualization effect when the data center model determines that the real face image is true. The attention map area of the real face has a small region range, while the last column shows that the data center model mistakenly classifies the real face with a different attention map. 
 \emph{We find that when the forgery detection model makes the wrong decision, the attention map would not focus on suitable local face regions};
In  Figure 8 (b) the first eight columns represent the visualization results when the data center model identifies the input fake face image as false. 
The attention map area of these inputs almost focuses on artifact regions (like mouth, eyes and nose).
However, the last column shows the attention map when the data center model mistakenly identifies the fake face as true. 
It can be found that these attention maps don't focus on the semantic regions, but instead on backgrounds;
In Figure 8 (c), the first eight columns represent the visualization results when the global server model identifies the real face image as true.
Similarly in Figure 8 (a), the proposed model mainly focuses on the local semantic regions. 
For the wrong identification as shown in the last column, the results show the model pays much attention on redundant face regions but not local regions;
In Figure 8 (d), the first eight columns represent the visualization results when the global server model identifies the forged face image as fake.
Similarly in Figure 8 (b), the model also extracts discriminative forgery representation in local semantic artifact regions. 
The wrong results as shown in the last column show the model is mistaken by the disturbance of complex backgrounds.

 Overall, these visualization results clearly prove our proposed FedFogery could extract discriminative forgery representation to focus on local semantic face regions to improve interpretability.
 Additionally, it is interesting to find that the visualization attention map could help to directly identify the artifact regions to expand the forgery detection field in the future.

 \begin{figure}[ht] 
\centering
\subfloat[Data Center (Real)]{
\includegraphics[width=8.5cm]{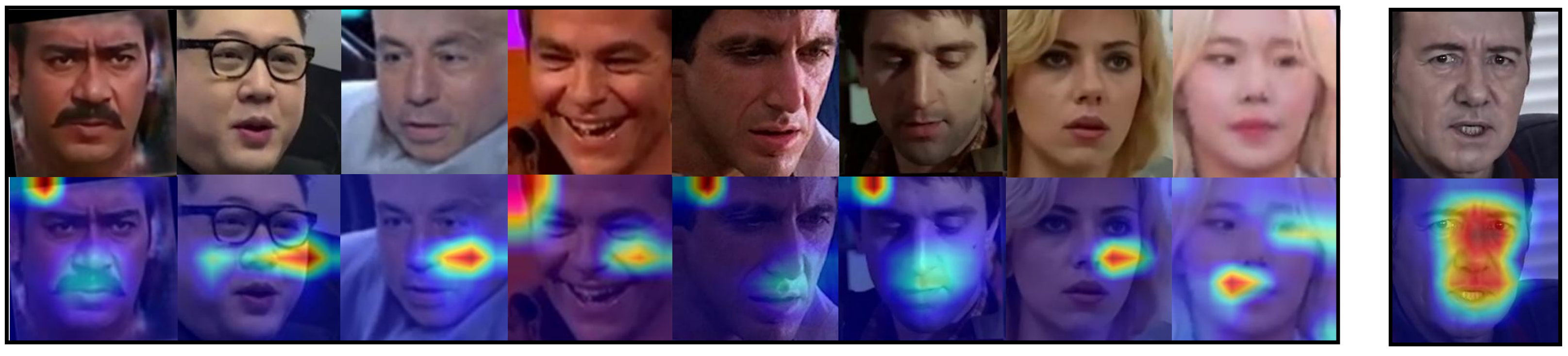}
}
\quad
\subfloat[Data Center (Fake)]{
\includegraphics[width=8.5cm]{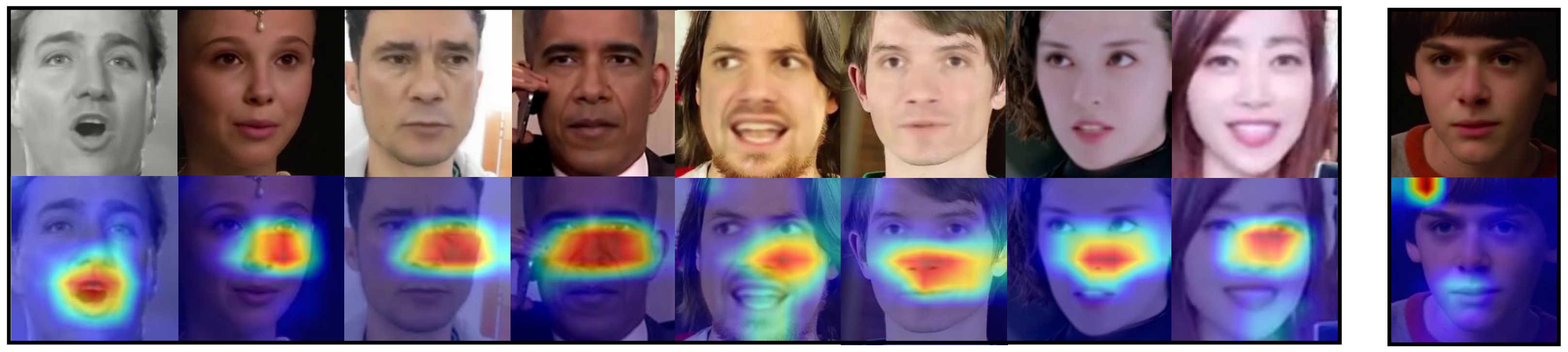}
}
\quad
\subfloat[Global Server (Real)]{
\includegraphics[width=8.5cm]{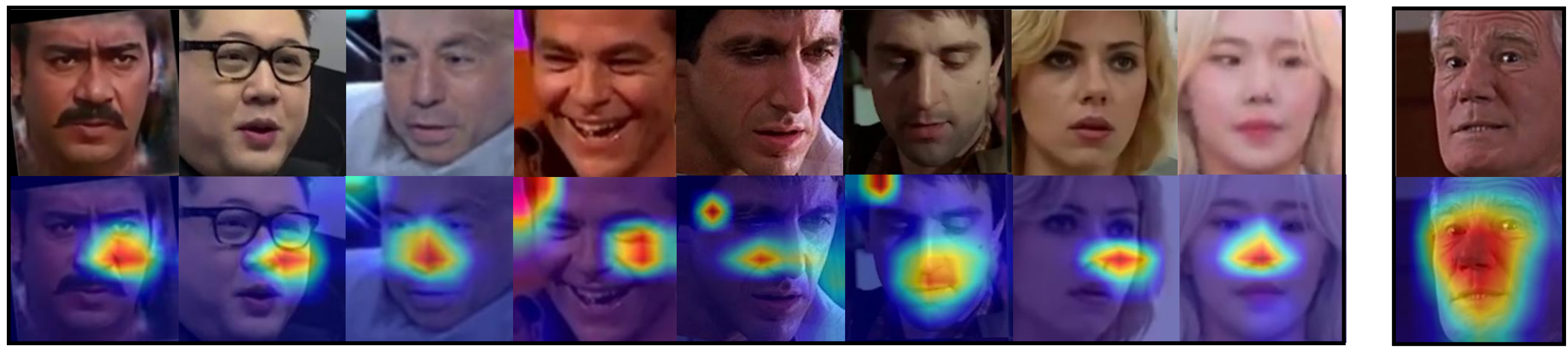}
}
\quad
\subfloat[Global Server (Fake)]{
\includegraphics[width=8.5cm]{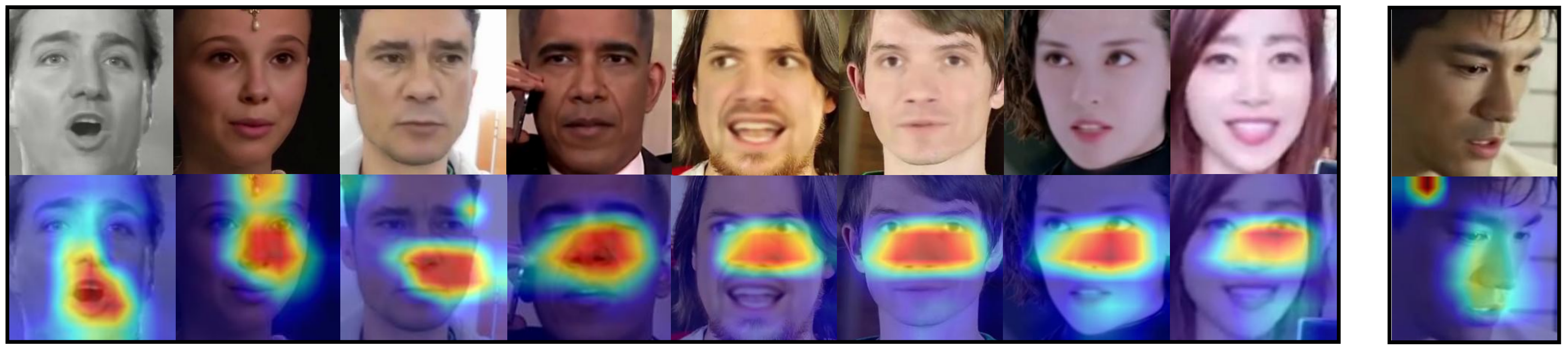}
}
\caption{\textbf{Grad-CAM attention map}. The first eight columns show the designed model identifies the input authenticity correctly; the last column shows the designed model makes wrong decisions.}
\end{figure}
\subsection{Future Directions}
Although data-driven forgery detection methods have achieved great progress in recent years, which benefit from the contraption of deep network architecture and large-scale labeled data. To further mimic real-world scenarios, more and more researchers focus on the generalization ability in face forgery detection tasks. Thus, we construct the novel hybrid-dataset and generalized-dataset forgery detection task to focus on detecting forgery clues in cross-domain even unknown artifact types. In this section, we discuss several unsolved problems and future potential directions to inspire researches in the community.}

\textbf{Unsupervised/Weakly Supervised Forgery Detection}:  Nowadays, most existing face forgery detection models are trained with fully supervised training data. However, more and more unknown artifact patterns would be created with the development of the deep generative model. And it is not impossible to label every artifact type accurately on the Internet. Thus, the key solution is to learn the general forgery clues to boost the forgery detection model performance. Unsupervised or weakly supervised learning(e.g. self-attention mechanism \cite{zhuang2022uia}, dynamic convolution\cite{zhao2018dynamic}, meta-learning \cite{zhao2020recognizing}) provides a possible solution to learn the essential discrepancy between real and fake images, which can strengthen the usability and scalability.

\textbf{Robust Forgery Detection Representation}: In real-world scenarios, these forgery detection models will be applied in complex scenarios, which results in image changes in illumination, occlusion, perturbation, etc. Meanwhile, malicious attack algorithms would possibly force the detection model to make the wrong decision. Thus, learning a robust detection representation to defend adversarial examples and noises is an important goal in security and privacy protection fields.

\textbf{Interpretable Forgery Detection}: Most existing deep learning-based model is still a “black box”, which lacks theoretical explanation. It is necessary and important to explore interpretability, not only high accuracy. The interpretable forgery detection could help expand more real application scenarios. Meta-learning and hybrid knowledge models may provide a feasible way to solve the interpretability problem in the future.

\section{Conclusions}
In this paper, we propose a novel generalized residual Federated learning method for face Forgery detection (FedForgery). 
To improve the forgery detection generalization, we design a variational autoencoder to learn robust discriminative residual feature maps to detect forgery clues. 
Additionally, the residual federated learning framework is introduced to train the forgery detection model collaboratively with multiple local data centers, which is essential to protect data privacy. 
We conduct several experiments for face forgery detection tasks in complex real-world scenarios.
The experimental results demonstrate that the proposed FedForgery could effectively improve the ability of model generalization and privacy protection. 
In the future, we will focus on exploring robust and interpretable face forgery detection methods to mimic more real-world scenarios. 


\bibliographystyle{IEEEtran}
\bibliography{ref.bib}
\end{document}